\documentclass[10pt,twocolumn,letterpaper]{article}

\usepackage{ijcb}
\usepackage{times}
\usepackage{epsfig}
\usepackage{graphicx}
\usepackage{amsmath}
\usepackage{amssymb}
\usepackage{multirow}
\usepackage[table,xcdraw]{xcolor}
\ijcbfinalcopy % *** Uncomment this line for the final submission

\ifijcbfinal\fi
\begin{document}

%%%%%%%%% TITLE
\title{TypeNet: Scaling up Keystroke Biometrics}

\author{Alejandro Acien, Aythami Morales, Ruben Vera-Rodriguez, Julian Fierrez\\
School of Engineering, Universidad Autonoma de Madrid, Spain\\
{\tt\small \{alejandro.acien, aythami.morales, ruben.vera, julian.fierrez\}@uam.es}
\and
John V. Monaco\\
Naval Postgraduate School, Monterrey CA, USA\\
{\tt\small vinnie.monaco@nps.edu}
}

\maketitle
\thispagestyle{empty}

%%%%%%%%% ABSTRACT
\begin{abstract}
   We study the suitability of keystroke dynamics to authenticate 100K users typing free-text. For this, we first analyze to what extent our method based on a Siamese Recurrent Neural Network (RNN) is able to authenticate users when the amount of data per user is scarce, a common scenario in free-text keystroke authentication. With 1K users for testing the network, a population size comparable to previous works, TypeNet obtains an equal error rate of 4.8\% using only 5 enrollment sequences and 1 test sequence per user with 50 keystrokes per sequence. Using the same amount of data per user, as the number of test users is scaled up to 100K, the performance in comparison to 1K decays relatively by less than 5\%, demonstrating the potential of TypeNet to scale well at large scale number of users. Our experiments are conducted with the Aalto University keystroke database. To the best of our knowledge, this is the largest free-text keystroke database captured with more than 136M keystrokes from 168K users.
\end{abstract}

%%%%%%%%% BODY TEXT
\section{Introduction}
Keystroke dynamics is a behavioral biometric trait aimed to recognize individuals based on their typing habits. The velocity of pressing and releasing different keys \cite{Banerjee}, the hand postures during typing \cite{Buschek}, and the pressure exerted when pressing a key \cite{10.1007/978-3-030-31321-0_2} are some of the features taken into account by keystroke biometric algorithms aimed to discriminate among users. Although keystroke technologies suffer of high intra-class variability, especially in free-text scenarios (i.e. the input text typed is not previously fixed), the ubiquity of keyboards as a method of text entry makes keystroke dynamics a near universal modality to authenticate users on the Internet.

Text entry is prevalent in day-to-day applications: unlocking a smartphone, accessing a bank account, chatting with acquaintances, email composition, posting content on a social network, and e-learning \cite{2020_AAAI_edBB_JH}. As a means of user authentication, keystroke dynamics is economical because it can be easily integrated into the existing computer security systems with minimal alteration and user intervention. These properties have prompted several companies to capture and analyze keystrokes. The global keystroke dynamics market will grow from \$$129.8$ million dollars to \$$754.9$ million by 2025, a rate of up to $25\%$ per year \cite{2019alliedmarket}. As an example, Google has recently committed \$7 million dollars to fund TypingDNA \cite{2019silicon}, a startup company which authenticates people based on their typing behavior.

At the same time, the security challenges that keystroke dynamics promises to solve are constantly evolving and getting more sophisticated every year: identity fraud, account takeover, sending unauthorized emails, and credit card fraud are some examples \cite{2019sec}. In this context, keystroke biometric algorithms capable of authenticating individuals while interacting with computer applications are more necessary than ever. However, these challenges are magnified when dealing with applications that have hundreds of thousands to millions of users.

The literature on keystroke biometrics is extensive, but to the best of our knowledge, these systems have only been evaluated with up to several hundred users. While other popular biometrics such as fingerprint and face recognition have been evaluated at the million-user scale \cite{Schroff}, the performance of keystroke biometrics in large scale scenarios remains unpublished.

The aim of this paper is to explore the feasibility and limits of scaling a free-text keystroke biometric authentication system to $100$,$000$ users. The main contributions of this work are threefold: 

\begin{enumerate}
\setlength\itemsep{0em}
\item We introduce TypeNet, a free-text keystroke biometrics system based on a Siamese Recurrent Neural Network (RNN) trained on $55$M keystrokes from $68$K users, suitable for user authentication at large scale.
\item We evaluate TypeNet in terms of Equal Error Rate (EER) as the number of test users is scaled from $100$ to $100$,$000$ (independent from the training data). TypeNet learns a feature representation of a keystroke sequence without need for retraining if new subjects are added to the database. Therefore, TypeNet is easily scalable.
\item We carry out a comparison with previous state-of-the-art approaches for free-text keystroke biometric authentication. The performance achieved by the proposed method outperforms previous approaches in the scenarios evaluated in this work. 

\end{enumerate}

In summary, we present the first evidence in the literature of competitive performance of free-text keystroke biometric authentication at large scale ($100$K test users). The results reported in this work demonstrate the potential of this behavioral biometric for widespread deployment.

The paper is organized as follows: Section \ref{related_works} summarizes related works in free-text keystroke dynamics to set the background. Section \ref{aalto} describes the dataset used for training and testing TypeNet. Section \ref{system_description} describes the processing steps and learning methods in TypeNet. Section \ref{experimental_protocol} details the experimental protocol. Section \ref{experiments_results} reports the experiments and analyze the results obtained. Section \ref{conclusions} summarizes the conclusions and future work.

%-------------------------------------------------------------------------
\section{Related Works and Background}
\label{related_works}

Keystroke biometric systems are commonly placed into two categories: \textit{fixed-text}, where the keystroke sequence typed by the user is prefixed, such as a username or password, and \textit{free-text}, where the keystroke sequence is arbitrary, such as writing an email or transcribing a sentence with typing errors, and different between training testing. Biometric authentication algorithms based on keystroke dynamics for desktop and laptop keyboards have been predominantly studied in fixed-text scenarios where accuracies higher than $95\%$ are common \cite{2016_IEEEAccess_KBOC_Aythami}. Approaches based on sample alignment (e.g. Dynamic Time Warping) \cite{2016_IEEEAccess_KBOC_Aythami}, Manhattan distances \cite{Vinnie1}, digraphs \cite{Bergadano}, and statistical models (e.g. Hidden Markov Models) \cite{Ali} have shown to achieve the best results in fixed-text.

Nevertheless, the performances of free-text algorithms are generally far from those reached in the fixed-text scenario, where the complexity and variability of the text entry contribute to intra-subject variations in behavior, challenging the ability to recognize users \cite{Sim}. Monrose and Rubin \cite{Monrose} proposed in 1997 a free-text keystroke algorithm based on user profiling by using the mean latency and standard deviation of digraphs and computing the Euclidean distance between each test sample and the reference profile. Their results worsened from $90\%$ to $23\%$ of correct classification rates when they changed both user's profiles and test samples from fixed-text to free-text. Gunetti and Picardi \cite{Gunetti} extended the previous algorithm to n-graphs. They calculated the duration of n-graphs common between training and testing and defined a distance function based on the duration and order of such n-graphs. Their results of $7.33\%$ classification error outperformed previous state-of-the-art. Nevertheless, their algorithm needs long keystroke sequences (between $700$ and $900$ keystrokes) and many keystroke sequences (up to $14$) to build the user's profile, which limits the usability of that approach. Murphy \etal~\cite{Murphy} more recently collected a very large free-text keystroke dataset ($\sim$$2.9$M keystrokes) and applied the Gunetti and Picardi algorithm achieving $10.36\%$ classification error using sequences of $1$,$000$ keystrokes and $10$ genuine sequences to authenticate users.

More recently than the pioneering works of Monrose and Gunetti, some algorithms based on statistical models have shown to work very well with free-text, like the POHMM (Partially Observable Hidden Markov Models) \cite{Monaco}. This algorithm is an extension of the traditional Hidden Markov Models (HMMs), but with the difference that each hidden state is conditioned on an independent Markov chain. This algorithm is motivated by the idea that keystroke timings depend both on past events and the particular key that was pressed. Performance achieved using this approach in free-text is close to fixed-text, but it again requires several hundred keystrokes and has only been evaluated with a database containing less than 100 users.

Nowadays, with the proliferation of machine learning algorithms capable of analysing and learning human behaviors from large scale datasets, the performance of keystroke dynamics in the free-text scenario has been boosted. As an example, \cite{Ceker} proposes a combination of the existing digraphs method for feature extraction plus a Support Vector Machine (SVM) classifier to authenticate users. This approach achieves almost $0\%$ error rate using samples containing $500$ keystrokes. These results are very promising, even though it is evaluated using a small dataset with only $34$ users. More recently, in \cite{Deb} the authors employ a Recurrent Neural Network (RNN) within a Siamese architecture to authenticate users based on $8$ biometric modalities on smartphone devices. They achieved results in free-text of $81.61\%$ TAR (True Acceptance Rate) at $0.1\%$ FAR (False Acceptance Rate) using just $3$ second test windows with a dataset of $37$ users.

Previous works in free-text keystroke dynamics have achieved promising results with up to several hundred users (see Table \ref{table:works}), but they have yet to scale beyond this limit and leverage emerging machine learning techniques that benefit from vast amounts of data. Here we take a step forward in this direction of machine learning-based free-text keystroke biometrics by using the largest dataset published to date with $136$M keystrokes from $168$K users. We analyze to what extent deep learning models are able to scale in keystroke biometrics to authenticate users at large scale while attempting to minimize the amount of data per user required for enrollment.

\begin{table}
\begin{center}
\begin{tabular}{|l|c|c|c|c|}
\hline
Year [Ref] & \#Users  & \#Seq.  & Sequence Size & \#Keys\\
\hline\hline
1997 \cite{Monrose} & $31$	& N/A         & N/A             	  & N/A\\
2005 \cite{Gunetti} & $205$	& $1-15$ &	$700-900$ keys &	$688$K \\
2016 \cite{Ceker}   & $34$	& $2$         &	$\sim$ $7$ keys       & $442$K \\
2017 \cite{Murphy}  & $103$	& N/A     &	$1$,$000$ keys	            & $12.9$M \\
2018 \cite{Monaco}  & $55$	& $6$         &	$500$ keys	            & $165$K \\
2019 \cite{Deb}     & $37$	& $180$K      &	$3$ seconds	            & $6.7$M \\
\textbf{2020 Ours}           & \boldmath$168$K	& \boldmath$15$        &	 \boldmath $\sim$ $70$ \textbf{keys}& \boldmath$136$\textbf{M}\\
\hline
\end{tabular}
\end{center}
\caption{Comparison among different free-text keystroke datasets employed in relevant related works. N/A = Not Available.}
\label{table:works}
\end{table}

%-------------------------------------------------------------------------

\section{Keystroke Dataset}
\label{aalto}
All experiments are conducted with the Aalto University Dataset \cite{Dhakal} that comprises more than $5$GB of keystroke data collected from $168$,$000$ participants during a three month time span.  The acquisition task required subjects to memorize English sentences and then type them as quickly and accurate as they could. The English sentences were selected randomly from a set of $1$,$525$ examples taken from the Enron mobile email and gigaword newswire corpora. The example sentences contained a minimum of $3$ words and a maximum of $70$ characters. Note that the sentences typed by the participants could contain even more than $70$ characters because each participant could forget or add new characters when typing.

For the data acquisition, the authors launched an online application that records the keystroke data from participants who visit their webpage and agree to complete the acquisition task (i.e. the data was collected in an uncontrolled environment). Press (keydown) and release (keyup) event timings were recorded in the browser with millisecond resolution using the JavaScript function \texttt{Date.now}. All participants in the database completed $15$ sessions (i.e. one sentence for each session) on either a physical desktop or laptop keyboard. The authors also reported demographic statistics: $72$\% of the participants took a typing course, $218$ countries were involved, and $85$\% of the participants have English as native language.

\section{System Description}
\label{system_description}
\subsection{Pre-processing and Feature Extraction}
The raw data captured in each user session includes a time series with three dimensions: the keycodes, press times, and release times of the keystroke sequence. Timestamps are in UTC format with millisecond resolution, and the keycodes are integers between $0$ and $255$ according to the ASCII code.

\begin{figure}[t!]
\centering
\includegraphics[width=\columnwidth]{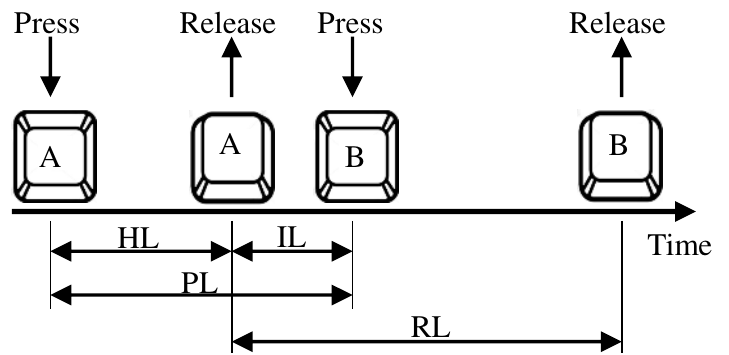}
\caption{Example of the 4 temporal features extracted between two consecutive keys: Hold Latency (HL), Inter-key Latency (IL), Press Latency (PL) and Release Latency (RL).}
\label{features}
\end{figure}

We extract $4$ temporal features for each sequence (see Figure \ref{features} for details): (i) Hold Latency (HL): the elapsed time between press and release key events; (ii) Inter-key Latency (IL): the elapsed time between releasing a key and pressing the next key; (iii) Press Latency (PL): the elapsed time between two consecutive press events; and Release Latency (RL): the elapsed time between two consecutive release events. These $4$ features are commonly used in both fixed-text and free-text keystroke systems \cite{Alsultan}. Finally, we include the keycodes as an additional feature.

The $5$ features are calculated for each keystroke in the sequence. Let $N$ be the length of the keystroke sequence, such that each sequence provided as input to the model is a time series with shape $N \times 5$ ($N$ keystrokes by $5$ features).
All feature values are normalized before being provided as input to the model. Normalization is important so that the activation values of neurons in the input layer of the network do not saturate (i.e. all close to $1$). The keycodes are normalized to between $0$ and $1$ by dividing each keycode by $255$, and the $4$ timing features are converted to seconds. This scales most timing features to between $0$ and $1$ as the average typing rate over the entire dataset is $5.1$ $\pm$ $2.1$ keys per second. Only latency features that occur either during very slow typing or long pauses exceed a value of $1$.
\subsection{The Deep Model: LSTM Architecture}
\begin{figure}[t!]
\centering
\includegraphics[width=\columnwidth]{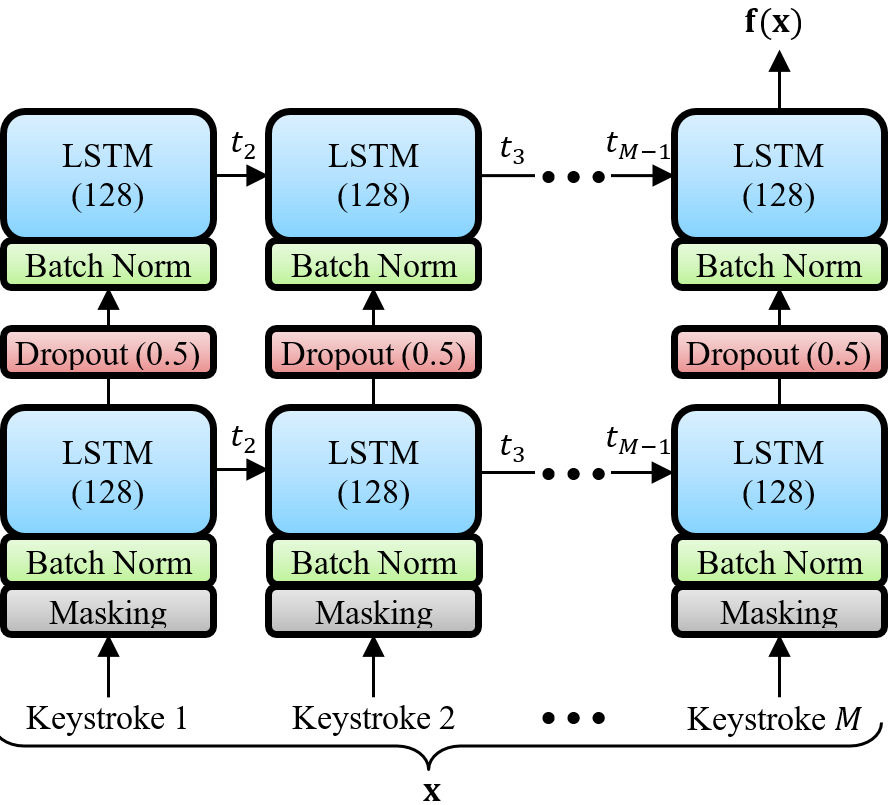}
\caption{Architecture of TypeNet for free-text keystroke sequences. The input \textbf{x} is a time series with shape $M \times 5$ (keystrokes $\times$ keystroke features) and the output $\textbf{f}($\textbf{x}$)$ is an embedding vector with shape $1\times128$.}
\label{LSTM}
\end{figure}
In keystroke dynamics, it is thought that idiosyncratic behaviors that enable authentication are characterized by the relationship between consecutives key press and release events (e.g. temporal patterns, typing rhythms, pauses, typing errors). In a free-text scenario, keystroke sequences may differ in both length and content. This reason motivates us to choose a Recurrent Neural Network (RNN) as our keystroke authentication algorithm. RNNs have demonstrated to be one of the best algorithms to deal with temporal data (e.g. \cite{2020_TIFS_BioTouchPass2_Tolosana}, \cite{Tolosana}) and are well suited for free-text keystroke sequences (e.g. \cite{Deb}, \cite{Lu}).

Our RNN model is depicted in Figure \ref{LSTM}. It is composed of two Long Short-Term Memory (LSTM) layers of $128$ units. Between the LSTM layers, we perform batch normalization and dropout at a rate of $0.5$ to avoid overfitting. Additionally, each LSTM layer has a dropout rate of $0.2$. 

One constraint when training a RNN using standard backpropagation through time applied to a batch of sequences is that the number of elements in the time dimension (i.e. number of keystrokes) must be the same for all sequences. Let's fix the size of the time dimension to $M$. In order to train the model with sequences of different lengths $N$ within a single batch, we truncate the end of the input sequence when $N>M$ and zero pad at the end when $N<M$, in both cases to the fixed size $M$. Error gradients are not computed for those zeros and do not contribute to the loss function at the output layer thanks to the Masking layer indicated in Figure~\ref{LSTM}.

Finally, the output of the model $\textbf{f}($\textbf{x}$)$ is an array of size $1 \times 128$ that we will employ later as an embedding feature vector to authenticate users.

%-------------------------------------------------------------------------
\section{Experimental Protocol}
\label{experimental_protocol}
Our goal is to build a keystroke biometric system capable of generalizing to new users not seen during model training. For this, we train our deep model in a Siamese framework which allows us to employ different users to train and test the authentication system. The RNN must be trained only once on an independent set of users. This model then acts as a feature extractor that provides input to a simple distance-threshold based authentication scheme. After training the RNN once, we evaluate authentication performance for a varying number of users and enrollment samples per user.
\subsection{Siamese Training}
In Siamese training, the model has two inputs (i.e. two keystroke sequences from either the same or different users), and therefore, two outputs (i.e. embedding vectors). During the training phase, the model will learn discriminative information from the pairs of keystroke sequences and transform this information into an embedding space where the embedding vectors (the outputs of the model) will be close in case both keystroke inputs belong to the same user (genuine pairs), and far in the opposite case (impostor pairs).

%\begin{figure*}[t!]
%\centering
%\includegraphics[scale=0.35]{Figures/Siamese.png}
%\caption{Architecture for a) training and b) testing the proposed TypeNet system. In training, a pair of keystroke feature sequences ($\textbf{x}_{i}$, $\textbf{x}_{j}$) together with their label $L_{ij}$ (which indicates if they are genuine or impostor comparisons) feed the Siamese setup and generate two embedding vectors $\textbf{f}(\textbf{x}_{i})$ and $\textbf{f}(\textbf{x}_{j})$. In testing, we compare $G$ genuine gallery feature vectors $\textbf{x}_{g}$ with a query $\textbf{x}_{q}$ by computing the Euclidean distances between the embedding vectors ($G$ one-to-one distances between each gallery and the test) and then averaging to obtain the final distance measure.}
%\label{Siamese}
%\end{figure*}

For this, we use the \textit{Contrastive loss} function defined specifically for this task \cite{Taigman}. Let $\textbf{x}_{i}$ and $\textbf{x}_{j}$ each be a keystroke sequence that together form a pair which is provided as input to the model. The contrastive loss calculates the Euclidean distance between the model outputs:
\begin{equation}
\label{disance}
     d_C(\textbf{x}_{i},\textbf{x}_{j})= \left \| \textbf{f}(\textbf{x}_{i}) - \textbf{f}(\textbf{x}_{j})\right \|
\end{equation}
where $\textbf{f}(\textbf{x}_{i})$ and $\textbf{f}(\textbf{x}_{j})$ are the model outputs (embedding vectors) for the inputs $\textbf{x}_{i}$ and $\textbf{x}_{j}$, respectively. The model will learn to make this distance small (close to $0$) when the input pair is genuine and large (close to $\alpha$) for impostor pairs by computing the loss function $\mathcal{L}$ defined as follows:
\begin{equation}
\label{loss}
     \mathcal{L}= (1-L_{ij})\frac{d^2(\textbf{x}_{i},\textbf{x}_{j})}{2}+L_{ij}\frac{\max^2\left \{0, \alpha-d(\textbf{x}_{i},\textbf{x}_{j})\right \} }{2}
\end{equation}
where $L_{ij}$ is the label associated with each pair that is set to $0$ for genuine pairs and $1$ for impostor ones, and $\alpha \geq 0$ is the margin (the maximum margin between genuine and impostor distances).

We train the RNN using only the first $68$K users in the dataset. From this subset we generate genuine and impostor pairs using all the $15$ keystroke sequences available for each user. This provides us with $15\times68$K$\times15=15.3$M impostor pair combinations and $15\times14/2=105$ genuine pair combinations for each user. The pairs were chosen randomly in each training batch ensuring that the number of genuine and impostor pairs remains balanced ($512$ pairs in total in each batch including impostor and genuine pairs). Note that the remaining $100$K users will be employed only to test the model, so there is no data overlap between the two groups of users (open-set authentication paradigm).

Regarding the training details, the best results were achieved with a learning rate of $0.05$, Adam optimizer with $\beta_{1} = 0.9$, $\beta_{2} = 0.999$ and $\epsilon = 10^{-8}$, and the margin set to $\alpha = 1.5$. The model was trained for $200$ epochs with 150 batches per epoch and $512$ sequences in each batch. The model was built in \texttt{Keras-Tensorflow}.

\subsection{Testing}
We authenticate users by comparing gallery samples $\textbf{x}_{g}$ belonging to one of the users in the test set to a query sample  $\textbf{x}_{q}$ from either the same user (genuine match) or another user (impostor match). The test score is computed by averaging the Euclidean distances $d_E$ between each gallery embedding vector $\textbf{f}(\textbf{x}_{g})$ and the query embedding vector $\textbf{f}(\textbf{x}_{q})$ as follows:
\begin{equation}
\label{score}
     \textit{score}= \frac{1}{G}\sum_{g=1}^{G} d_E(\textbf{f}(\textbf{x}_{g}),\textbf{f}(\textbf{x}_{q}))
\end{equation}
where $G$ is the number of sequences in the gallery (i.e. the number of enrollment samples). Taking into account that each user has a total of $15$ sequences, we retain $5$ sequences per user as test set (i.e. each user has $5$ genuine test scores) and let $G$ vary between $1 \leq G \leq 10$ in order to evaluate the performance as a function of number of enrollment sequences.

To generate impostor scores, for each enrolled user we choose one test sample from each remaining user. We define $K$ as the number of enrolled users. In our experiments, we vary $K$ in the range $100 \leq K \leq 100$,$000$. Therefore each user has $5$ genuine scores and $K-1$ impostor scores. Note that we have more impostor scores than genuine ones, a common scenario in keystroke dynamics authentication. The results reported in the next section are computed in terms of Equal Error Rate (EER), which is the value where False Acceptance Rate (FAR, proportion of impostors classified as genuine) and False Rejection Rate (FRR, proportion of genuine users classified as impostors) are equal. The error rates are calculated for each user and then averaged over all $K$ users \cite{2014_IWSB_Aythami_Keystroking}.

%-------------------------------------------------------------------------
\setlength{\tabcolsep}{8pt}
\renewcommand{\arraystretch}{1.6}
\begin{table}
\begin{center}
\begin{tabular}{cc|c|c|c|c|c|}
\cline{3-7}
\multicolumn{2}{c}{} &\multicolumn{5}{c|}{\cellcolor[HTML]{C0C0C0}\textbf{\#enrollment sequences per user $G$}} \\ 
\cline{3-7} 
\multicolumn{2}{c}{\multirow{-2}{*}{}} & \cellcolor[HTML]{C0C0C0}\textbf{1} & \cellcolor[HTML]{C0C0C0}\textbf{2} & \cellcolor[HTML]{C0C0C0}\textbf{5} & \cellcolor[HTML]{C0C0C0}\textbf{7} & \cellcolor[HTML]{C0C0C0}\textbf{10} \\ \hline
\multicolumn{1}{|c|}{\cellcolor[HTML]{C0C0C0}} & \cellcolor[HTML]{C0C0C0}\textbf{30}  & 9.53  & 8.00  & 6.43 & 5.95   & 5.49 \\
\cline{2-7} 
\multicolumn{1}{|c|}{\cellcolor[HTML]{C0C0C0}}  & \cellcolor[HTML]{C0C0C0}\textbf{50}  & 7.56                               & 6.04                               & 4.80                               & 4.23                               & 3.73                               \\ \cline{2-7} 
\multicolumn{1}{|c|}{\cellcolor[HTML]{C0C0C0}}                         & \cellcolor[HTML]{C0C0C0}\textbf{70}  & 7.06                               & 5.55                               & 4.38                               & 3.87                               & 3.35                               \\ \cline{2-7} 
\multicolumn{1}{|c|}{\cellcolor[HTML]{C0C0C0}}                         & \cellcolor[HTML]{C0C0C0}\textbf{100} & 6.98                               & 5.49                               & 4.29                               & 3.85                               & 3.33                               \\ \cline{2-7} 
\multicolumn{1}{|c|}{\multirow{-5}{*}{\cellcolor[HTML]{C0C0C0}\rotatebox{90}{\textbf{\#keys per sequence $M$}}}} & \cellcolor[HTML]{C0C0C0}\textbf{150} & 6.97                               & 5.46                               & 4.29                               & 3.85                               & 3.33                               \\ \hline
\end{tabular}
\end{center}
\caption{Equal Error Rate ($\%$) achieved for different values of the parameters $M$ (sequence length) and $G$ (number of enrollment sequences per user).}
\label{table:performance}
\end{table}
%-------------------------------------------------------------------------
\section{Experiments and Results}
\label{experiments_results}
\subsection{Performance vs User Data}
As commented in the related works section, one key factor when analyzing the performance of a free-text keystroke authentication algorithm is the amount of keystroke data per user employed for enrollment. In this work, we study this factor with two variables: the keystroke sequence length $M$ and the number of gallery sequences used for enrollment $G$.

Our first experiment reveals to what extent $M$ and $G$ affect the authentication performance of our model. Note that the input to our model has a fixed size of $M$ after the Masking process shown in Figure~\ref{LSTM}. For this experiment, we set $K = 1$,$000$ where $K$ is the number of enrolled users.

Table \ref{table:performance} summarizes the error rates achieved for the different values of sequence length $M$ and enrollment sequences per user $G$. We can observe that for sequences longer than $M = 70$ there is no significant improvement in the performance. Adding three times more key events (from $M = 50$ to $M = 150$) lowers the EER by only $0.57\%$ for all values of $G$. However, adding more sequences to the gallery shows greater improvements with about $50\%$ relative error reduction when going from $1$ to $10$ sequences independent of $M$. The best results are achieved for $M = 70$ and $G = 10$ with an error rate of $3.35\%$. For one-shot authentication ($G = 1$), our approach has an error rate of $7.06\%$ using sequences of $70$ keystrokes. These results suggest that our approach achieves a performance close to that of a fixed-text scenario (within $\sim$$5\%$ error rate) even when the data is scarce. For the following experiments, we set $M = 50$ and $G = 5$ to have a good trade-off between performance and amount of user data.

\subsection{Comparison with State-of-the-Art Works}
We now compare the proposed TypeNet with our implementation of two state-of-the-art algorithms for free-text keystroke authentication: one based on statistical models, the POHMM (Partially Observable Hidden Markov Models) from \cite{Monaco}, and another algorithm based on digraphs and SVM from \cite{Ceker}. To allow fair comparisons, all models are trained and tested with the same data and experimental protocol: $G = 5$ enrollment sequences per user, $M = 50$ keystrokes per sequence, $K = 1$,$000$ test users.
\begin{figure}[t!]
\centering
\includegraphics[width=\columnwidth]{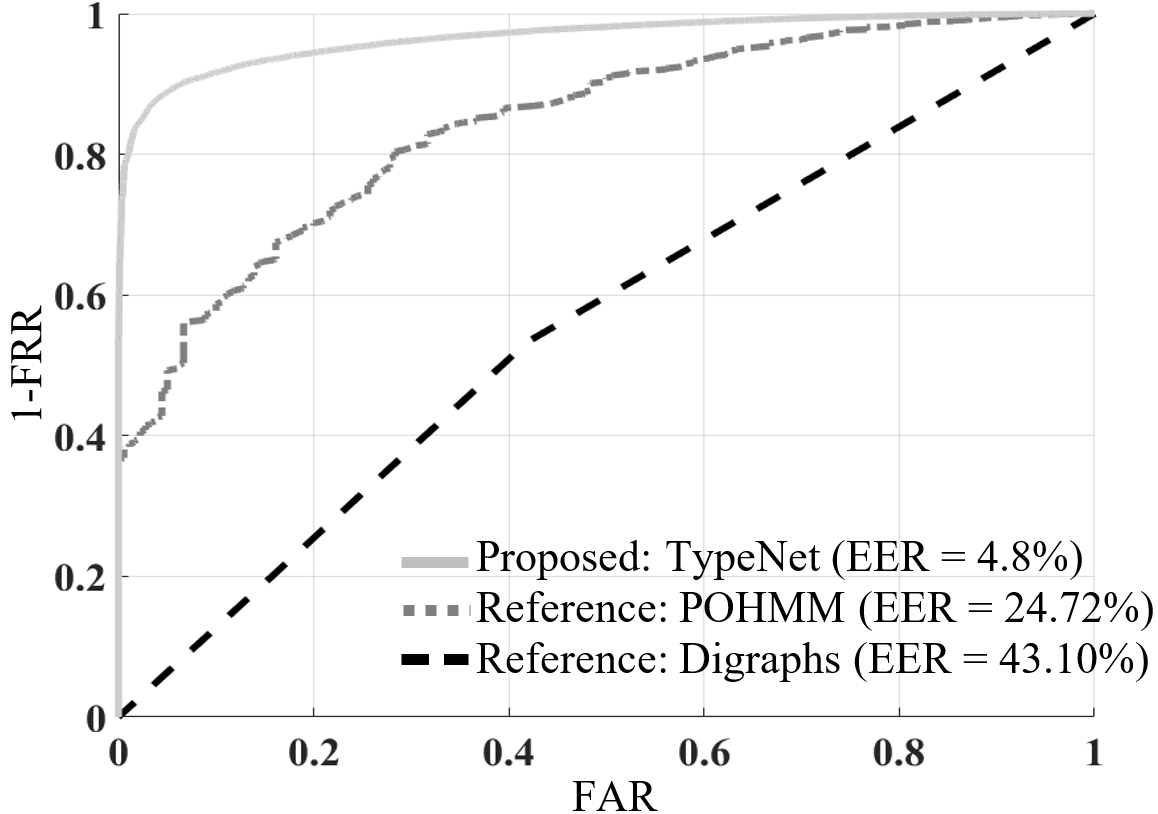}
\caption{ROC comparison in free-text biometric authentication between the proposed TypeNet and two state-of-the-art approaches: POHMM from \cite{Monaco} and digraphs/SVM from \cite{Ceker}. $M = 50$ keystrokes per sequence, $G = 5$ enrollment sequences per user, 1 test sequence per user, and $K = 1$,$000$ test users.}
\label{ROCs}
\end{figure}

In Figure \ref{ROCs} we plot the performance of the three approaches with the Aalto dataset described in Section~\ref{aalto}. We can observe that TypeNet outperforms previous state-of-the-art free-text algorithms in this scenario where the amount of enrollment data is reduced ($5 \times M=250$ training keystrokes in comparison to more than $10$,$000$ in related works, see Section~\ref{related_works}), thanks to the Siamese training step. The Siamese RNN has learned to extract meaningful features from the training dataset, which minimizes the amount of data needed for enrollment. The SVM generally requires a large number of training sequences per user ($\sim$$100$), whereas in this experiment we have only $5$ training sequences per user. We hypothesize that the lack of training samples contributes to the poor performance (near chance accuracy) of the SVM.

\subsection{User Authentication at Large Scale}
In the last experiment, we evaluate to what extent our model is able to generalize without performance decay. For this, we scale the number of enrolled users $K$ from $100$ to $100$,$000$. Remember that for each user we have $5$ genuine test scores and $K - 1$ impostor scores, one against each other test user. The model used for this experiment is the same trained for previous section ($68$,$000$ independent users included in the training phase). 
\begin{figure}[t!]
\centering
\includegraphics[width=\columnwidth]{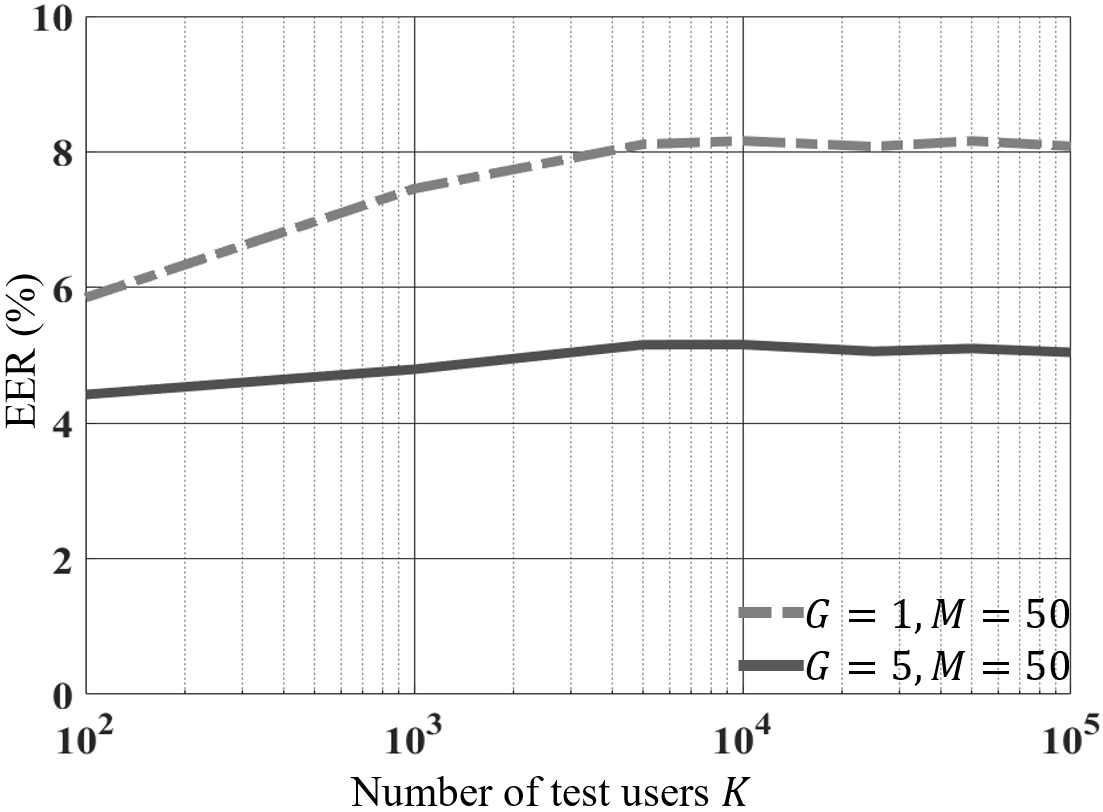}
\caption{EER of our proposed TypeNet when scaling up the number of test users $K$ in one-shot ($G = 1$ enrollment sequences per user) and balanced ($G = 5$) authentication scenarios. $M = 50$ keystrokes per sequence.}
\label{EERs}
\end{figure}

Figure \ref{EERs} shows the authentication results for one-shot enrollment ($G = 1$ enrollment sequences, $M = 50$ keystrokes per sequennce) and the balanced scenario ($G = 5$, $M = 50$) for different values of $K$. We can observe that for both scenarios there is a slight performance decay when we scale from $100$ to $5$,$000$ test users, which is more pronounced in the one-shot scenario. However, for a large number of users ($K \geq 10$,$000$), performance stabilizes in both scenarios. These results demonstrate the potential of the Siamese RNN architecture in TypeNet to authenticate users at large scale in free-text keystroke dynamics.
%-------------------------------------------------------------------------
\section{Conclusions}
\label{conclusions}
We have presented TypeNet, a new free-text keystroke biometrics system based on a Siamese RNN architecture, and experimented with it at large scale in a dataset of $136$M keystrokes from $168$K users. Siamese networks have shown to be effective in face recognition tasks when scaling up to hundreds of thousands of identities. The same capacity has been also shown by our TypeNet in free-text keystroke biometrics. 

In all scenarios evaluated, specially when there are many users but few enrollment samples per user, the results achieved in this work suggest that our model outperforms previous state-of-the-art algorithms. Our results range from $9.53\%$ to $3.33\%$ EER, depending on the amount of user data enrolled. A good balance between performance and the amount of enrollment data per user is achieved with $5$ enrollment sequences and $50$ keystrokes per sequence, which yields an EER of $4.80\%$ for $1$K test users. Scaling up the number of test users does not significantly affect the performance: the EER of TypeNet decays only $5\%$ in relative terms with respect to the previous $4.80\%$ when scaling up from $1$K to $100$K test users. Evidence of the EER stabilizing around $10$K users demonstrates the potential of this architecture to perform well at large scale.
	
For future work, we will improve the way training pairs are chosen in Siamese training. Recent work has shown that choosing \textit{hard pairs} during the training phase can improve the quality of the embedding feature vectors \cite{Wu}. We plan to test our model with other databases, and investigate smarter ways to combine the multiple sources of information \cite{2018_INFFUS_MCSreview2_Fierrez}, e.g., the multiple distances in Equation~(\ref{score}).

\section{Acknowledgment}
This work has been supported by projects: PRIMA (MSCA-ITN-2019-860315), TRESPASS (MSCA-ITN-2019-860813), BIBECA (RTI2018-101248-B-I00 MINECO), and by edBB (UAM). A. Acien is supported by a FPI fellowship from the Spanish MINECO.

{\small
\bibliographystyle{ieee}
\bibliography{main}
}

\end{document}